%% file: main.tex
\documentclass[unnumsec,webpdf,contemporary,large]{oup-authoring-template}%
\usepackage{url}
\usepackage{graphicx}
\usepackage{listings}
\usepackage{verbatim}
\usepackage{xcolor}
\usepackage{amsmath}
\usepackage{hyperref}
\usepackage{algorithm}
\usepackage{natbib}
\usepackage{authblk}
\usepackage{footnote}
\begin{document}
\journaltitle{Journal Title Here}
\DOI{DOI HERE}
\copyrightyear{2022}
\pubyear{2019}
\access{Advance Access Publication Date: Day Month Year}
\appnotes{Paper}
\title[]{Data-Driven Information Extraction and Enrichment of Molecular Profiling Data for Cancer Cell Lines}

% \author[1]{Ellery Smith}
% \author[2,3]{Rahel Paloots}
% \author[4]{Dimitrios Giagkos}
% \author[2,3]{Michael Baudis}
% \author[1]{Kurt Stockinger}
% \affil[1]{Zurich University of Applied Sciences, Switzerland}
% \affil[2]{University of Zurich, Switzerland}
% \affil[3]{Swiss Institute of Bioinformatics, Switzerland}
% \affil[4]{Infili Technologies, Greece}

% List of authors, with corresponding author marked by asterisk
\author{Ellery Smith$^{\ast,1}$, Rahel Paloots$^{\ast,2,3}$, Dimitris Giagkos$^4$, Michael Baudis$^{2,3}$, Kurt Stockinger$^1$}
% Author addresses
\affil{$^1$Zurich University of Applied Sciences, Switzerland, $^2$University of Zurich, Switzerland, $^3$Swiss Institute of Bioinformatics, Switzerland, $^4$Infili Technologies, Greece}
% \\[2pt]
% E-mail address for correspondence
% {smil@zhaw.ch}
% {rahel.paloots@uzh.ch}

\received{Date}{0}{Year}
\revised{Date}{0}{Year}
\accepted{Date}{0}{Year}

% Running headers of paper:
\markboth%
% First field is the short list of authors
{Smith, Paloots et al.}
% Second field is the short title of the paper
{Data Enrichment for Cancer Cell Lines}

% Add a footnote for the corresponding author if one has been
% identified in the author list

\abstract{
\textbf{Motivation}: With the proliferation of research means and computational methodologies, published biomedical literature is growing exponentially in numbers and volume~(\citealp{LUBOWITZ20213221}). Cancer cell lines are frequently used models in biological and medical research that are currently applied for a wide range of purposes, from studies of cellular mechanisms to drug development, which has led to a wealth of related data and publications. Sifting through large quantities of text to gather relevant information on the cell lines of interest is tedious and extremely slow when performed by humans.  Hence, novel computational information extraction and correlation mechanisms are required to boost meaningful knowledge extraction.
\textbf{Results}: In this work, we present the design, implementation and application of a novel data extraction and exploration system. This system extracts deep semantic relations between textual entities from scientific literature to enrich existing structured clinical data in the domain of cancer cell lines. We introduce a new public data exploration portal, which enables automatic linking of genomic copy number variants plots with ranked, related entities such as affected genes. Each relation is accompanied by literature-derived evidences, allowing for deep, yet rapid, literature search, using existing structured data as a springboard.
\textbf{Availability and Implementation}: Our system is publicly available on the web at \url{https://cancercelllines.org}.
\textbf{Contact}: The authors can be contacted at ellery.smith@zhaw.ch or rahel.paloots@uzh.ch.
}
\keywords{Cancer cell lines, copy number variants, natural language processing, information extraction}

\maketitle
\footnotetext{$^\ast$To whom correspondence should be addressed.}

\input{sections/00-introduction.tex}
\input{sections/01-methods_materials.tex}

\input{sections/02-results.tex}

\input{sections/03-discussion.tex}
%\input{sections/04-software.tex}
%\input{sections/05-conclusions.tex}

%\section{Supplementary Material}
%\label{sec:supplementary}

\section*{Acknowledgments}
This project has received funding from the European Union’s Horizon 2020 research and innovation program under grant agreement No 863410. MB receives support from the ELIXIR European bioinformatics organization for work related to the development of the GA4GH beacon protocol.

{\it Conflict of Interest}: None declared.

\section*{Data Availablity}

The data underlying this article are available at \url{https://github.com/progenetix/cancercelllines-web} and \url{https://pubmed.ncbi.nlm.nih.gov/}.

\bibliographystyle{plainnat}
\bibliography{refs}

\end{document}

%% file: sections/00-introduction.tex
\section{Introduction}
\label{sec:introduction}

%{\DG{With the proliferation of research means and computational methodologies, published biomedical  literature is growing exponentially in numbers and volume~\cite{LUBOWITZ20213221}. As a consequence, in the fields of biological, medical and clinical research the domain experts have to sift through massive amounts of scientific text to find relevant information. However, the process is extremely tedious and slow when performed by humans. Hence, novel computational information extraction and correlation mechanisms are required  to boost meaningful knowledge extraction.}

Cancer research is one of the most challenging and promising biomedical areas as reflected in the amount of attention it receives~(\citealp{Elmore2021-ft, Cabral2018-mz, Siegel2022-rb}). Cancer cell lines are important models for the study of cancer-related  pathophysiological mechanisms as well as for  pharmacological development and testing procedures. Cell lines are obtained from patient-derived malignant tissue and are cultivated \textit{in vitro}, potentially in an ``immortal'' way. Cancer cell lines are supposed to retain most of the genetic properties of the originating cancer (\citealp{Mirabelli2019-wv}), including genomic modifications that are characteristic for the respective disease's pathology and are absent in normal tissues. 

A class of  mutations ubiquitous in primary tumors and derived cell lines are  genomic \textit{copy number variants} (CNVs) which represent structural genome variations in which genomic segments of varying sizes have been duplicated or deleted from one or both alleles. The set of CNVs observed in a given tumor (``CNV profile'') frequently includes one or multiple changes characteristic for a given tumor type. For instance, while many colorectal carcinomas display duplications of chromosome 13 (\citealp{lassmanCNVinCRC2007, baudis_5918_2007}), neuroepithelial tumors frequently show small, often bi-allelic deletions involving the CDKN2A gene locus on the short arm of chromosome 9 (\citealp{cdkn2ameningiomas, hoischen_cgh_2008, raoGlioblastomas2008}). Recurring CNV events are supposed to be driven by their selective advantage for cancer cells, \textit{i.e.} recurrently duplicated regions predominately will affect genes favorable for a clonal expansion (``oncogenes'') and, conversely, deleted regions will frequently contain growth-limiting (``tumor-suppressor'') genes (\citealp{vogelstein2013cance}).

The collection and comparative analysis of cancer and cancer cell line CNV data is important for the understanding of disease mechanisms as well as the discovery of potential therapeutics. Progenetix (\citealp{baudis_progenetix_2001, progenetix2021}) is a knowledge resource for oncogenomic variants, mainly focusing on representing cancer CNVs. A recent spin-off from the  Progenetix resource is \textit{cancercelllines.org} - a database dedicated to genomic variations in  cancer cell lines. In addition to CNVs, \textit{cancercelllines.org} also includes information about sequence variations such as single nucleotide variants (SNVs), assembled from the aggregation of genomic analysis data of cell line instances. Currently over 16,000 cell lines from over 400 different cancer diagnoses are represented in this resource. %To enable hierarchical representation of cancers, National Cancer Institute's "NCIt" classification codes are used (\citealp{ncithesaurus}).

%Currently it contains information for 142,063 cancer related samples, representing more than 800 different cancer diagnoses according to the National Cancer Institute's "NCIt"  hierarchical cancer classification codes (\citealp{ncithesaurus}). Among the Progenetix data over 5,000 CNV profiles represent instances of approximately 2000 different cancer cell lines.

%TODO - main motivating example
Natural language processing (NLP) has proven to be a game-changer in the field of clinical information processing for attaining pivotal knowledge in the healthcare domain (\citealp{10.1007/s10115-022-01779-1}). In fact, numerous studies have been undertaken in exploring indirect relations between drugs, diseases, proteins and genes from unstructured text provided in literature resources. One among many is (\citealp{Subramanian}), where the authors systematically design an NLP pipeline for drug re-purposing via evidence extraction from PubMed abstracts. Even though such studies exhibit some promising performance, neither ground truth is considered for further relevance evaluation of discovered drug-cancer therapeutic associations, nor visualization of results is provided. Additionally, SimText (\citealp{10.1093/bioinformatics/btab365}), a text mining toolset built for visualization of similarities among biomedical entities, manages to extract and display knowledge interconnections from user-selected literature text. However, no quantitative metrics were presented for evaluating the efficiency of the utilized NLP methods.

%While the Progenetix collection represents the largest public resource of curated cancer CNV data, in its current state the data content is largely static and focused on the representation of sample-specific and aggregated CNV profiles as well as a limited amout of sample-associated biomedical and technical annotations such as diagnostic classifications and provenance data. 

In this paper we study how to use state-of-the-art information extraction algorithms such as LILLIE (\citealp{SMITH2022101938}) to identify known mutated genes and find out which genes are most likely affected in certain CNV regions. As a result, \textbf{we introduce a novel data exploration system, allowing for the dynamic visualization and exploration of previously orthogonal data models by extracting and enriching information from both structured and unstructured data}.An overview of the architecture of our system is shown in Figure~\ref{fig:systemarchitecture}. By using the tool, the user will be able to visualize gene information extracted by our algorithm on the CNV profiles of cancer cell lines. The source code for our system is available to the public on GitHub\footnote{\url{https://github.com/progenetix/cancercelllines-web}}.

%% file: sections/01-methods_materials.tex
\section{Methods and Materials}
\label{sec:methods}

\begin{figure*}[b]
    \centering
    \includegraphics[scale=0.5]{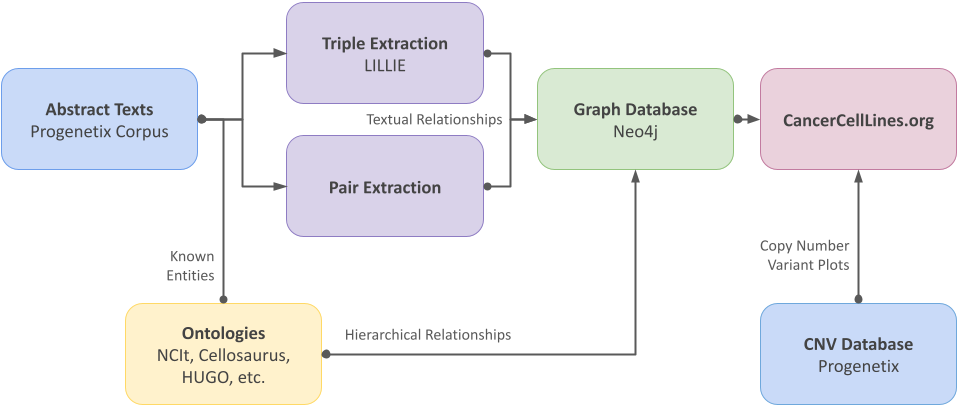}
    \caption{An overview of the architecture of our system, which provides a bridge between unstructured textual corpora, and structured clinical data. We first use abstract texts from the Progenetix corpus, along with entity names and synonyms from existing biomedical ontologies such as NCIt and Cellosaurus, to identify textual relational triples using the LILLIE Open Information Extraction system. We then use these triples, along with the relationships from these ontologies, to build a graph database, which is then mapped to existing Copy Number Variant plots from the Progenetix structured database.}
    \label{fig:systemarchitecture}
\end{figure*}

\subsection{Proposed Method}
The Progenetix project curates individual cancer CNA (Copy Number Aberration) profiles and associated metadata from published oncogenomic studies and data repositories, which, over the past 22 years, has resulted in the most comprehensive representation of cancer genome CNA profiling data available today (\citealp{progenetix2021}). The project consists of both these structured CNA profiles and a manually-curated corpus of the associated literature from which these data were derived, but currently the two remain as heterogeneous entities from an automated exploration standpoint. In this paper we propose a novel end-to-end methodology that aims to bridge this gap by combining information extracted from unstructured text (i.e., publication abstracts from PubMed) with structured knowledge resources (i.e., Progenetix and cancercelllines.org) in order to construct an interface for exploratory analysis of positionally mapped genomic variations based on literature evidence.

Our work mainly consists of two parts: i. fine-tuning LILLIE (\citealp{SMITH2022101938}), a state-of-the-art information extraction tool, in the cancer cell lines context and ii. development of a portal that serves as the interface for linking various genomic CNV findings with evidence extracted from literature text. 

\begin{figure*}[h]
    \centering
    \includegraphics[scale=0.6]{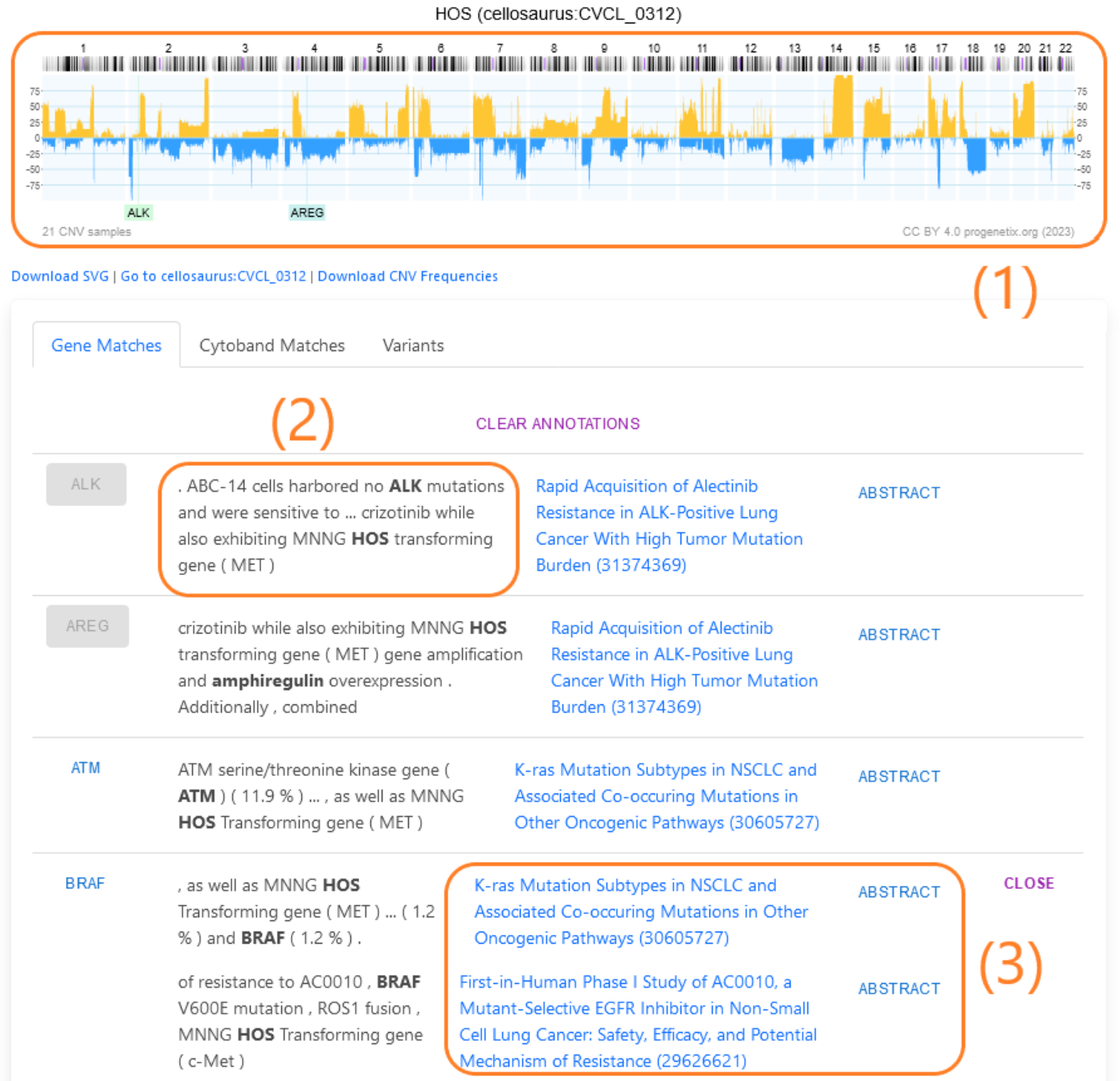}
    \caption{ A sample of the results available for the cell line \textit{HOS}, including: (1) associated genomic locations mapped on the copy number variation profile plot (gain CNVs yellow, loss CNVs in blue); (2) evidences for each result; (3) and the relevant abstracts from which the results were derived. The results columns are, from left to right: Gene, Cytoband, or other entity labels; Primary evidence for each abstract (the relevant cell line/entity annotations are marked in bold); Abstract title, and a link to the corresponding PubMed article; Expand/Collapse controls to view detailed information (shown in Figure~\ref{fig:helaevidence}).}
    \label{fig:webresultshos}
\end{figure*}

More specifically, we use cell lines as a jumping-off point to provide our literature extraction results. For each cell line, we visualize a corresponding CNV plot, which is annotated by selected extracted genes, and a categorized, ranked list of related entities, as shown in Figure~\ref{fig:webresultshos} and on the results page of our system\footnote{\url{https://cancercelllines.org/cellline/?id=cellosaurus:CVCL_0312}}. We provide the most relevant evidence for the given result alongside the title of each paper, allowing the user to easily check the validity of the result, and a toggle to expand each result, revealing the full annotated abstract text, as shown in Figure~\ref{fig:helaevidence}.

\subsection{Information Extraction from Unstructured Text}

While there are many existing systems which focus on either the topic of biomedical text extraction (\citealp{10.1007/s10115-022-01779-1}) or the creation of knowledge graphs from text (\citealp{knowledge-extraction-cohort-characteristics, 71be196e376b4fa1a71a1bbda69a8d4e, 10.1093/bioinformatics/btab851}), the \textit{main challenge of our approach was to merge these two concepts} with an existing structured database, such that both can be explored in parallel, and provide complementary information in a streamlined fashion.

Rather than using a known benchmarking dataset for either information extraction or knowledge graph creation, as for example explored in \citealp{10.1093/bioinformatics/btz600}, we designed our system using an existing live knowledge base, with a focus on pragmatic data exploration of real-world data, rather than test-set performance. 

\begin{figure}[h]
    \centering
    \includegraphics[scale=0.3]{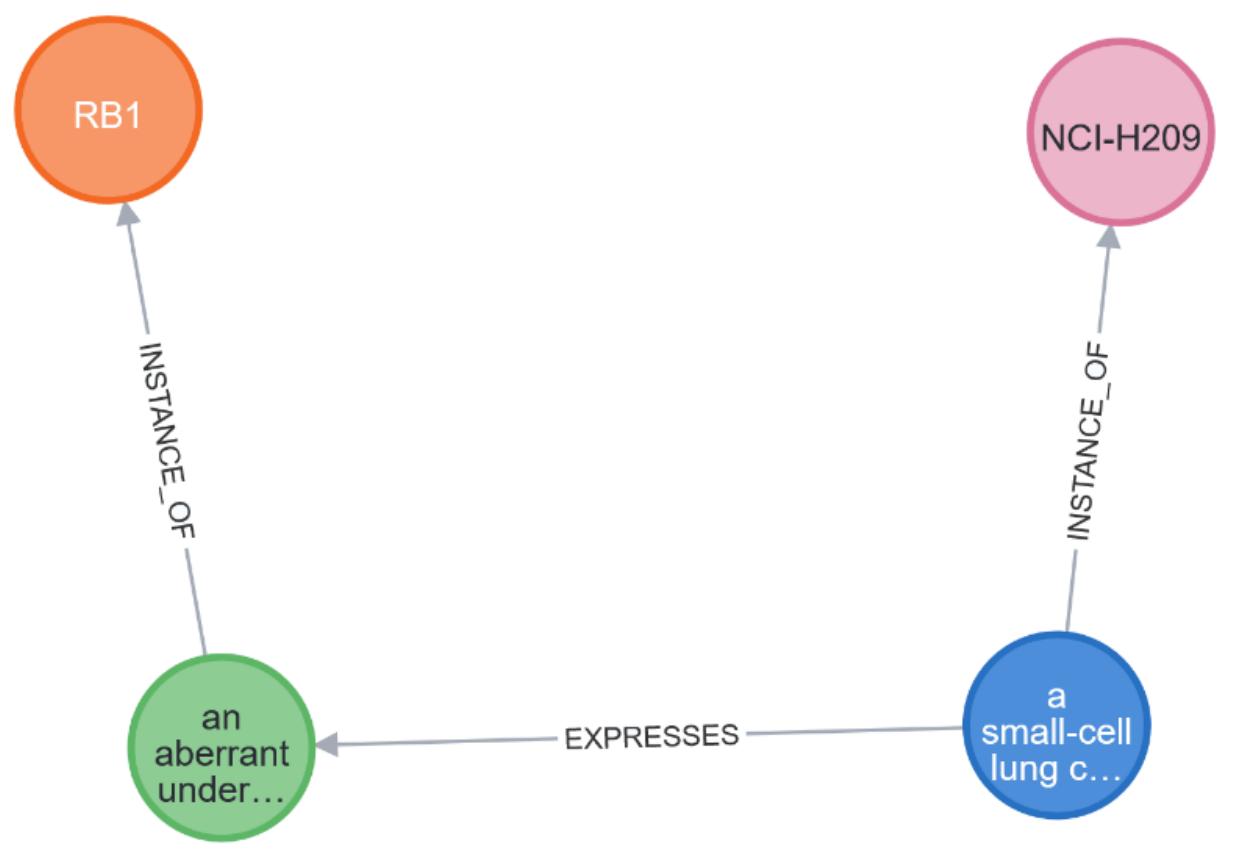}
    \caption{Graph representation of the relationships in the text ``\textit{a small-cell lung cancer cell line (NCI-H209) expresses an aberrant underphosphorylated form of the retinoblastoma protein RB1}'', deriving an \textsc{EXPRESSES} relationship between the cell line \textit{NCI-H209} and the gene \textit{RB1}.}
    \label{fig:relationexample}
\end{figure}

Prior work in the development of the LILLIE system (\citealp{SMITH2022101938}), provides a test-case for the concept of applying open information extraction to database enrichment. When given a piece of unstructured text, subject-predicate-object relational triples are output by LILLIE. The format of these triples is as follows: given the sentence ``A small-cell lung cancer cell line (NCI-H209) expresses an aberrant underphosphorylated form of the retinoblastoma protein RB1.'', the following triple is output:

\begin{verbatim}
    (RB1 [the retinoblastoma protein RB1] ;
     EXPRESSES [expresses an aberrant
                underphosphorylated form] ;
     NCI-H209 [small-cell lung cancer
               cell line (NCI-H209)]) 
\end{verbatim}

Where the subject, \textit{RB1}, predicate, \textit{EXPRESSES} and object, \textit{NCI-H209}, are annotated by their textual context, shown in square brackets. The use of named entities in combination with textual context allows for entities to be matched to structured data, while providing additional contextual information alongside each connection. An example of this is shown in Figure~\ref{fig:relationexample}.

While the system was designed for the purposes of Open Information Extraction (OpenIE), which works across a broad range of textual domains, the potential for the application of the same system to a closed-domain task is also demonstrated, along with the viability of the output triples to be integrated to a structured database.

The LILLIE system gives state-of-the-art performance on the most widely-used OpenIE benchmarking datasets, CaRB and Re-OIE16, with F1 scores of 66.4\% and 53.9\% respectively, versus 61.7\% and 52.7\% for OpenIE6 (\citealp{openie6}) and 61.7\% and 53.5\% for IMoJIE (\citealp{imojie}), and AUC score improvements of up to 6\% over the previous state-of-the-art. Additionally, it includes features for adjusting and fine-tuning the output triples to potentially match any closed-domain task. 

The system consists of a \emph{pre-trained learning-based component}, based on open-domain corpora, and a \emph{hand-optimized rule-based component}, which is parameterized to allow fine-tuning of the output rules for a closed-domain task. In \citealp{SMITH2022101938}, it is shown that the adjustment of these rules can produce a customized Information Extractor for the use-case of linking anatomical entities and diseases with gene expressions, which can then be integrated into a relational database.

Here, we expand upon this work by applying a customized version of LILLIE to extract relational triples in the context of cancer cell lines, and integrate these triples into a graph database, allowing existing hierarchical ontologies to be linked with textual relationships in a structured manner, which we describe in detail in the following sections.

\subsubsection{Triple Extraction}

Other recent work in the field of biomedical information extraction focuses on recognizing a predecided set of relations (\citealp{knowledge-discovery-reuse-pipeline,biored}). However, here we use the open information extraction paradigm (\citealp{liu2016learning}) to extract any potential relationship between entities in the text, namely, as natural language subject-predicate-object triples. The use of this model allows a researcher to explore richer and more descriptive relations between entities than if they were mapped to discrete categories, and takes into account the fact that relationships in oncogenomics are often complex and subtle. Thus, open information extraction methods, coupled with domain expertise, was determined to be optimal for this use case. 

For this work, we use a customized version of the LILLIE system, for which we tailor the rules in the rule-based component to better suit the context of this task. Since scientific abstracts are typically written in a formal manner, many sentences contain complex conjunctions, and long-distance dependencies over multiple clauses. However, the language used is unambiguous and well-constructed, unlike in open-domain tasks. For example, the sentence ``2-O-Methylmagnolol Upregulates the Long Non-Coding RNA, GAS5, and Enhances Apoptosis in Skin Cancer Cells'' is complex to parse, but unambiguous in meaning.

As described in \citealp{SMITH2022101938}, it is possible to modify the triple parser and the output format of the LILLIE system to suit the target linguistic profile of the input texts. It is shown that in the context of biomedical paper abstracts, enabling and disabling certain features can produce an increase in entity-matched triples of 46.5\% over the unmodified version of LILLIE used in the open-domain context, and an increase of 42\% and 41\% over OpenIE6 (\citealp{openie6}) and IMoJIE (\citealp{imojie}), respectively, in the same context. In addition, we found that in this context, due to these linguistic factors, modifying LILLIE to use only the rule-based component produced a higher quality of results based on qualitative analysis of the end-user exploration portal by domain experts.

\subsection{Entity Linking}

After running the LILLIE system on the abstracts of all research articles in the Progenetix corpus to produce a set of textual triples for each abstract. Next, we match the subject and object with their corresponding entities in the following ontologies:

\begin{itemize}
    \item The cancer section of the NCIt thesaurus (\citealp{ncithesaurus})
    \item The UBERON anatomical ontology (\citealp{uberon})
    \item The Cellosaurus cell-line index (\citealp{cellosaurus})
    \item Cytogenetic mapping information from Progenetix (\citealp{progenetix2021})
    \item The HUGO gene nomenclature (\citealp{hugogene})
\end{itemize}

Each ontology provides an identifier for a given entity, along with a canonical name and a set of synonyms for each entity. We use dictionary-based methods (\citealp{quimbaya2016named}) to match a triple to its corresponding entities. Firstly, we pre-process the entity names in each ontology. For gene and cell line canonical names, we leave these in their original state without any processing, as even changing the case of a gene (e.g. ``STEP'' and ``step'') can cause ambiguities. Otherwise, we expand each of the synonyms provided by the ontologies into two forms: one which has been lemmatized and tokenized, and another which has only been case-normalized. We use these processed ontologies to construct a categorical entity dictionary.

We then represent the text of the subject and object of each triple in three forms: unprocessed, tokenized and lemmatized, and case-normalized. Using our dictionary, we mark each triple as containing an entity when a token overlap occurs with one of the entity forms in our dictionary. This method was chosen to emphasise precision and reduce spurious matches, since in biomedical text extraction, particularly when the domain is narrow, dictionary-based approaches for entity-relation extraction have been shown to give comparable performance to learning-based methods (\citealp{10.1007/s10115-022-01779-1}), due to the fact that textual biomedical entity references are typically precise, unambiguous and correspond directly with their canonical names.

\begin{figure}[h]
    \centering
    \includegraphics[scale=0.3]{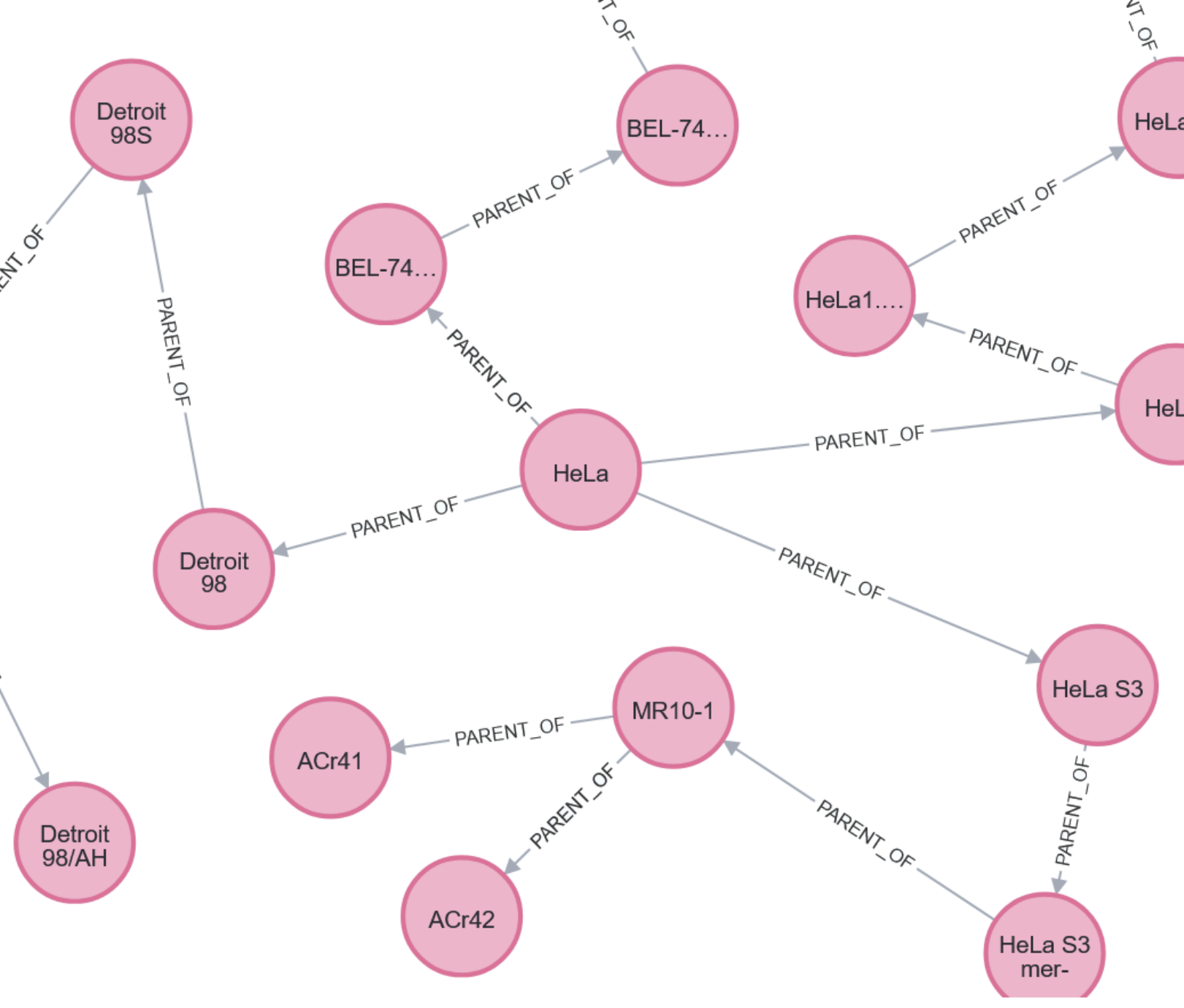}
    \caption{Portion of the cell line hierarchy for \textit{HeLa}, showing the entity itself, and its daughter cell lines. Nodes in the graph are derived from the ontologies (in this case, Cellosaurus), and the edges indicate a `\textit{parent-of}' relationship.}
    \label{fig:graphontologysample}
\end{figure}

In addition to an entity dictionary, we also use the ontologies described above to construct a hierarchical graph ontology, where entities are nodes, and parent-child relationships are represented by edges in the graph. A sample of this ontology is shown in Figure~\ref{fig:graphontologysample}, which depicts a portion of the resultant subgraph for the cell line \textit{HeLa}, derived from Cellosaurus.

We then place these entity-matched triples in a graph database, as shown in Figure~\ref{fig:relationexample}. Unlike in other works on biomedical knowledge graph building (\citealp{10.1093/bioinformatics/btz600}), we do not infer any relations using the graph itself. Only relations directly implied by the text are present in the graph, as shown in Figure~\ref{fig:relationexample}. By contrast, \citealp{10.1093/bioinformatics/btz600} attempt to synthesize whether a relationship exists in between two nodes based on existing relationships using a knowledge graph embeddings approach. Our aim here is to provide a link between existing evidences (between natural language and structured data), rather than synthesize new knowledge using machine learning methods.

\subsubsection{Pair Extraction}

\begin{figure*}[h]
    \centering
    \includegraphics[scale=0.42]{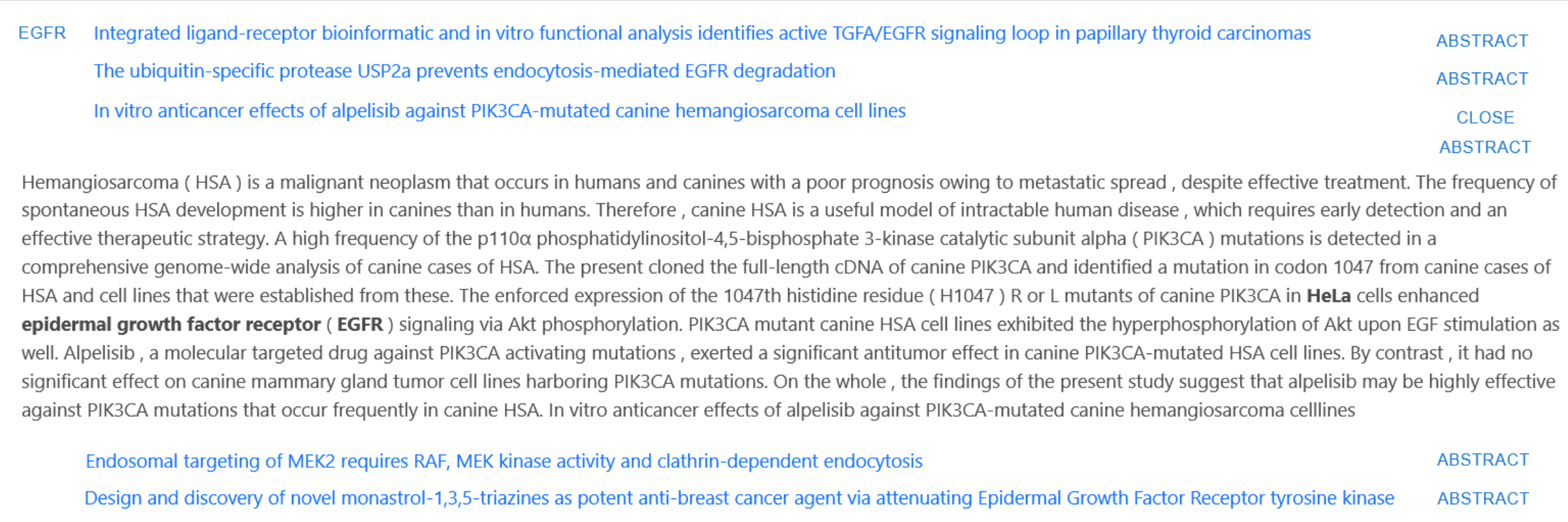}
    \caption{Section of the results demonstrating a relationship between the cell line \textit{HeLa} and the gene \textit{EGFR}, showing the paper title, primary evidence for the result, and, when expanded, the full annotated abstract text.}
    \label{fig:helaevidence}
\end{figure*}

\begin{figure*}[h]
    \centering
    \includegraphics[scale=0.38]{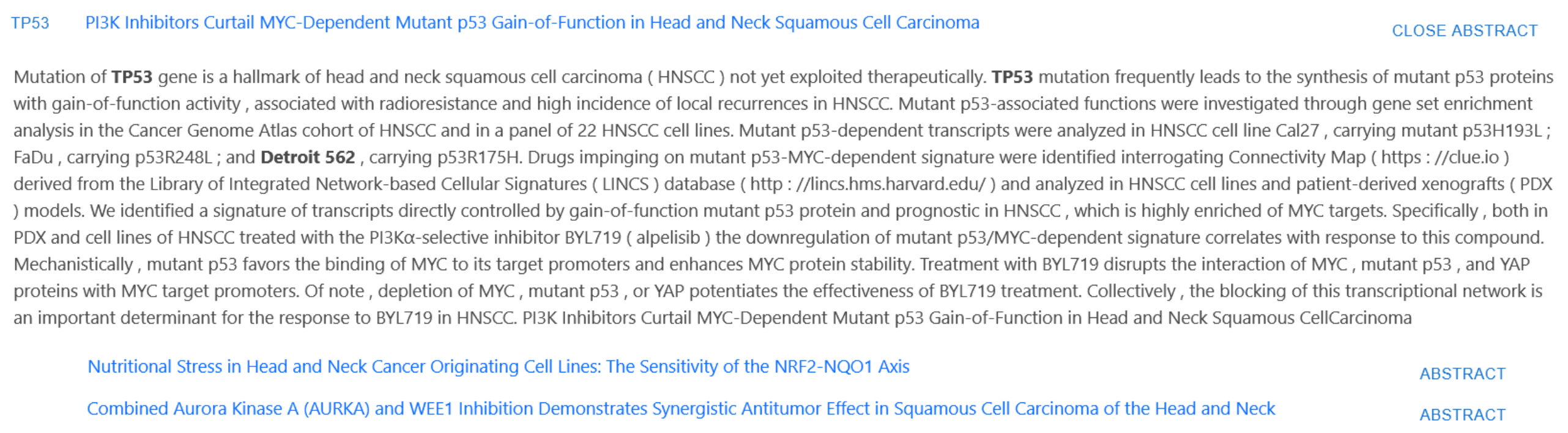}
    \caption{Section of the results demonstrating a relationship between the cell line \textit{Detroit 562} and the gene \textit{TP53}, showing the paper title, primary evidence for the result, and, when expanded, the full annotated abstract text.}
    \label{fig:detroitevidence}
\end{figure*}

While triple extraction can expose deep semantic relations between entities, this approach does not necessarily provide a complete representation of all relationships within the text, as it only extracts predicates that are directly expressed as singular verb phrases. An example of a strong sentential relationship extracted by triples is shown in the abstract in Figure~\ref{fig:helaevidence}, whereas long-distance relationships, as shown in Figure~\ref{fig:detroitevidence}, are not currently reliably extractable using similar methods. This is a known shortcoming of current information extraction techniques, and recent efforts such as BioRED (\citealp{biored}) have attempted to mitigate this deficiency by providing a corpus of long-distance relations that may span an entire document. However, the BioRED corpus is limited in both the number of relationship annotations and the fact that no specific annotations for cell lines are provided.

Naturally, if two entities are present in the same textual snippet, they are likely related in some manner, though this is not easily represented in the standard subject-predicate-object model. As such, we augment our triple extraction with what we term as \textit{pair extraction}, where we extract subject-object pairs, but leave the relation expressed as a simple numerical quantity. We combine these pair extractions with triples to produce a more representative ranking in the final output.

As such, we \textit{augment our high-precision triples with an additional high-recall method to capture long-distance relations}, using information retrieval techniques such as term distance. We use the following metric to rank relationships between entities:

$$
    R(D,e_1,e_2) = \sum_{P(e_1) \in D}\sum_{P(e_2) \in D}\log_2^{-1}\left(|P(e_1) - P(e_2)| + 1\right)
$$

Where $D$ is an abstract document, $e_1$ and $e_2$ are the pair of entities found in the abstract, $R(d,e_1,e_2)$ is the relationship score between $e_1$ and $e_2$ in a given abstract, and $P(e_n)$ is the token span position of entity $e_n$ in the abstract. Essentially, we model the potential relationship between entities by the sum of the inverse logarithms of the token distance between each instance of a given entity. If an entity pair is also marked as being part of a triple, we set $R(D,e_1,e_2) \mathrel{+}= 1$ for each occurrence of a triple, as this represents a definitive semantic connection. Thus, when combined with our semantic triple extraction, we can, given a pair of entities of interest, such as a gene and a cell line, produce a ranked list of abstracts, ordered by the strength of their relation, using this metric. In our portal, we then produce a ranked list of genes for each cell line based on the total score for that pair across all the literature in our database.

An example of such a pair extraction is shown in Figure~\ref{fig:detroitevidence}, where a complex relationship between \textit{Detroit 562} and \textit{TP53} is extracted as a pair relationship, but cannot be represented as a semantic triple using even current state-of-the-art information extraction techniques. We capture this relationship using our augmented approach, but such an abstract would have a lower weighting than one containing a triple-based relationship, due to it being a weaker inference. However, by highlighting a potential relationship to the user, a researcher can make a judgement on its relevance, which bridges the gap while current information extraction techniques are insufficiently mature enought to perform such tasks alone. Conversely, in Figure~\ref{fig:helaevidence}, the cell line \textit{HeLa} is explicitly linked with \textit{EGFR} through a triple relation, and is ranked higher due to the known presence of a strong semantic correlation.

%% file: sections/02-results.tex
\section{Results}
\label{sec:results}

In this section we will first apply our information extraction system for analyzing various cancer types. Afterwards we will evaluate the performance of our automatic information extraction pipelines. In particular, we want to address the following two research questions:

\begin{itemize}
    \item \emph{Research question 1: How well does our information extraction pipeline work for studying cancer cell lines and for exploring potentially new information? }
    \item \emph{Research question 2: What is the performance of our automatic information extraction algorithm for combining structured and unstructured data, i.e. from a database for cancer cell lines and research abstracts from PubMed?}
\end{itemize}

\subsection*{Example Use Cases}
%\DG{TODO: Merge biomedical results with CS results}

To validate the efficacy of our approach, we analyzed the results of our novel information extraction pipeline and how the extracted data corresponds to cell line CNV profiles. We will now illustrate how to analyze two different cancer types using our approach with the help of two example use cases.

\subsubsection{Head and Neck Squamous Cell Carcinomas - Cell Line Detroit 562}

\begin{figure*}[h]
    \centering
    \includegraphics[scale=0.50]{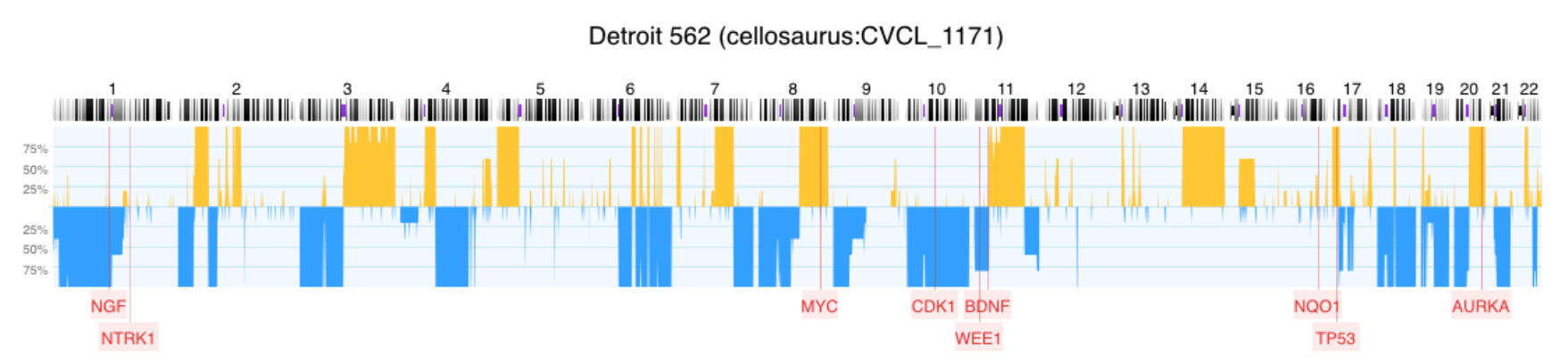}
    \caption{CNV frequency profile of 5 instances of the cancer cell line Detroit 562 annotated with enriched gene information from our information extraction pipeline. Copy number gains are shown in yellow and deletions in blue. Of note, a few of the regional CNVs deviate from the 100\% expected for stable clonal propagation due to some genomic instability and possible variation in the fidelity of individual profiling experiments. Mapping positions of genes of interest are shown in red.}
    \label{fig:detroit562}
\end{figure*}

Figure~\ref{fig:detroit562} depicts the CNV profile for Detroit 562 - a pharyngeal squamous cell carcinoma cell line (NCIT code C102872). Pharyngeal squamous cell carcinoma is a part of head and neck squamous cell carcinomas, often related to smokers. The results of our information extraction pipeline for genes AURKA and WEE1 claim that these genes are highly expressed and down-regulated\footnote{These results can be reconstructed here: \url{https://cancercelllines.org/cellline/?id=cellosaurus:CVCL_1171}} respectively in cancers, see (\citealp{lee2019combined}). This information is confirmed on the CNV profile where AURKA is duplicated and WEE1 is deleted. Similarly, MYC gene is brought forward as a possible target due to high expression and the region is duplicated on the CNV profile as well. 

Figure~\ref{fig:detroit562} also indicates TP53, a tumor-supressor gene involved in the control of cell division located on the short arm of chromosome 17. Due to its inhibitory role on cellular expansion, it is a frequent target of genomic deletions in a variety of cancers. However, TP53 can also acquire gain-of-function mutations that contribute e.g. to radio-resistance, thus explaining the duplication in this region in the case of a mutant allele (\citealp{ganci2020pi3k}). Conversely, NGF - a gene that is reported to be expressed in Detroit 562, exhibits alleleic deletion in our CNV data (\citealp{dudas2018nerve}), points towards alternative mechanisms responsible for its transcriptional activation.

\subsubsection{Breast Carcinomas - Cell Line MDA-MB-453}

\begin{figure*}[h]
    \centering
    \includegraphics[scale=0.50]{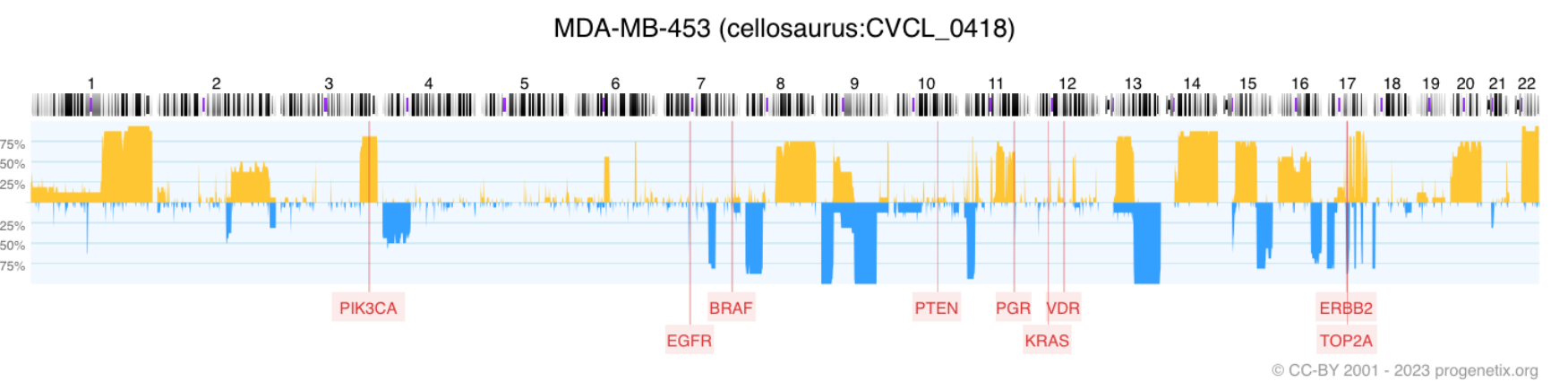}
    \caption{CNV frequency profile of 16 breast cancer cell line MDA-MB-453 samples, annotated with enriched gene information from our information extraction pipeline. Copy number gains are shown in yellow and deletions in blue. Mapping positions of genes of interest are shown in red.}
    \label{fig:MDA-MB-453}
\end{figure*}

Breast cancer is the most common cancer type in women, affecting more than 250,000 women in the US alone, see (\citealp{seer2017}). In breast cancer several clinico-pathological parameters have been recognized. One of the rare but clinically especially aggressive variants is the  ``triple-negative'' subtype, \textit{i.e.} where the tumor cells do not express 3 receptors commonly targeted in hormonal and immunotherapy: estrogen receptor, progesterone receptor and ERBB2 (HER2) receptor. Cell line MDA-MB-453 is a breast cancer cell commonly used to represent the triple-negative expression profile\footnote{\url{https://www.cellosaurus.org/CVCL\_0418}}. However, using our information extraction pipeline we could match this cell line to a publication that claimed its expression of ERBB2\footnote{These results can be reconstructed here: \url{https://cancercelllines.org/cellline/?id=cellosaurus:CVCL_0419}}, see (\citealp{santra2017identification}). Indeed, in our CNV data from 16 instances of MDA-MB-453 we can observe genomic duplications involving the ERBB2 locus on 17q (see Figure \ref{fig:MDA-MB-453}). 

While another paper claims PTEN to be expressed in MDA-MB-453 (\citealp{singh2011characterization}) the CNV profile does not indicate a genomic duplication event as causative and therefore indicating  transcriptional de-regulation. We also matched this cell line to 2 papers where mutation in KRAS was confirmed by our SNV data. Moreover, the expression of PIK3CA was confirmed by the duplication on the CNV profile as well as the mutation of the same gene was detected in the SNV data, see (\citealp{patra2017braf}).

Thus, to answer \textit{Research Question 1}: We show here that our novel information extraction feature facilitates further research into cancer cell lines. We were able to prove some known gene expression levels for cell lines Detroit 562 and for MDA-MB-453. Moreover, we could  discover some new or conflicting information about some other genes. 

More insights about how to reconstruct the exploration of these use cases with our system can be found at \url{https://docs.cancercelllines.org/literature-data/}.

\subsection*{Information Extraction}

After applying our information extraction pipeline for studying various cancer types, we will now evaluate the performance in terms of accuracy and number of relevantly-linked triples. 

\subsubsection{Data Exploration}

The Progenetix and cancercelllines.org resources provide PubMed identifiers for articles with a direct relation to genomic analyses in cancer cell lines. Crawling the PubMed database from these identifiers resulted in a corpus of 52,412 textual abstracts, which were used by our system to generate our graph database. As shown in Table~\ref{tbl:extractionstatistics}, we find 770,230 total entity matches, leading to a total of 12,139 distinct nodes in our graph.

\begin{table}[h]
\centering
\caption{Number of input abstracts, the number of matched entities found by our system, along with the number of cell lines extracted, and the number of unique relations per cell line.\\}
\def\arraystretch{1.1}
\begin{tabular}{|l|r|}
\hline
\textbf{Number of Abstracts}   & 52,412  \\ \hline
\textbf{Total Entity Matches} & 770,230 \\ \hline
\textbf{Unique Entity Matches} & 12,139  \\ \hline
\textbf{Unique Cell Lines} & 1,411 \\ \hline
\textbf{Abstracts per Cell Line} & 6.09 \\ \hline
\textbf{Linked Entities per Cell Line} & 53.609 \\ \hline
\end{tabular}
\label{tbl:extractionstatistics}
\end{table}

\subsubsection{Information Extraction Performance}

As there does not currently exist a benchmark for evaluating relationships between genes and cell lines, we adapt a subset of the BioRED benchmark (\citealp{biored}) to assess the performance of our system. The BioRED benchmark provides annotations for two types of evaluation: entity spans for NER (Named Entity Recognition) and entity-relationship pairs. For the NER task, annotations are provided for both Genes and Cell Lines, which fit the needs of this task. However, while this corpus includes relationship annotations for certain entity types (for example Gene-Gene and Gene-Chemical), it does not include Gene-CellLine relationships, which is the target of our platform.

To account for this, we adapt the BioRED annotations as follows: we assume that, for a given abstract, discussion of a given gene and cell line in the same passage implies a causal relationship between the two. While this may introduce false-positives, our exploration platform is designed to extract all potential relations between genes and cell lines, and then allows a domain expert to select those that they believe to be relevant. As such, this adaption of the test corpus represents a viable evaluation metric of the goals of our system, when combined with our qualitative analysis in Section~\ref{sec:results}.

Based on this, we generate a set of Gene-CellLine relationship pairs for each paper as follows: if both a gene and cell line are annotated as entities in the same abstract, we additionally annotate that paper with a relationship pair for these entities. We then perform the Entity Pair evaluation as described in \citealp{biored}. Our system achieves an F1 score of 74.2\% on our adapted BioRED test set for Gene-CellLine pairs. On the original BioRED corpus, BioBERT and PubMedBERT achieve scores between 58.8\% and 58.1\% respectively (for Chemical-Variant relationships) and 78.5\% and 78.1\% (for Gene-Gene relationships) on the Entity Pair test. We believe that these results, in combination with Table \ref{tbl:extractionstatistics} answers \textit{Research Question 2} posed at the beginning of this section.

% The average end-to-end time taken to add new abstracts to our database is $3.56$ seconds per paper, measured on a system with the following specifications: Intel Xeon W-11855M CPU @ 3.20GHz, 64GB RAM, NVIDIA RTX A4000 GPU. Our system includes a mechanism for adding new entries to the database from a provided list of PubMed IDs, and this time includes: crawling, information extraction, pair indexing and database construction (as described in Section~\ref{sec:discussion}).

%% file: sections/03-discussion.tex
\section{Discussion}
\label{sec:discussion}

Starting from a domain specific resource for curated genomic and associated data in cancer cell lines we extended its ``classical'' online database paradigm towards a knowledge exploration resource through the implementation of our novel literature information extraction algorithms. This change enables researchers to use the existing data - such as annotated genomic variations, visual indication of structural variation events and disease-related annotations - to gain context specific insights into molecular mechanisms through exploration of the added literature-derived information either directly or to prioritize follow-up analyses. We could show that the extracted results can easily be related to the resource's hallmark CNV profiles and this combination opens possibilities for knowledge expansion, including the critical evaluation of pre-existing annotations which may be affected by the fast-mutating nature of cancer cell lines. In our information extraction implementation we have shown that an interactive bimodal exploration model can be achieved in a streamlined manner, even if one data source comprises unstructured information.

Ubiquitous application of high-throughput molecular analyses as well as their interpretation in an ever increasing amount of publications drive a  ``data deluge'' in biomedical research. Our work demonstrates an application of information extraction techniques to add a knowledge exploration dimension to a genomic data resource. By doing so, we provide a tool to increase the speed and depth of scientific research using computational linguistic methods. 

For this work, we provide an evaluation on an adapted version of the BioRED corpus (and benchmarking of our information extractor is provided in \citealp{SMITH2022101938}). In addition, we provide a qualitative analysis in Section~\ref{sec:results}, demonstrating that our system can be applied on real-world data to discover new knowledge, and we envision our system as a tool to dynamically discover novel data in tandem with a domain expert.

%% file: main.bbl
\begin{thebibliography}{38}
\providecommand{\natexlab}[1]{#1}
\providecommand{\url}[1]{\texttt{#1}}
\expandafter\ifx\csname urlstyle\endcsname\relax
  \providecommand{\doi}[1]{doi: #1}\else
  \providecommand{\doi}{doi: \begingroup \urlstyle{rm}\Url}\fi

\bibitem[Bairoch(2018)]{cellosaurus}
Amos Bairoch.
\newblock The cellosaurus, a cell-line knowledge resource.
\newblock \emph{J. Biomol. Tech.}, 29\penalty0 (2):\penalty0 25--38, July 2018.

\bibitem[Baudis(2007)]{baudis_5918_2007}
M~Baudis.
\newblock Genomic imbalances in 5918 malignant epithelial tumors: an
  explorative meta-analysis of chromosomal cgh data.
\newblock \emph{BMC Cancer}, 7\penalty0 (1):\penalty0 226, 2007.

\bibitem[Baudis and Cleary(2001)]{baudis_progenetix_2001}
M~Baudis and ML~Cleary.
\newblock Progenetix.net: an online repository for molecular cytogenetic
  aberration data.
\newblock \emph{Bioinformatics}, 17\penalty0 (12):\penalty0 1228–1229, 2001.

\bibitem[Bostrom et~al.(2001)Bostrom, Meyer-Puttlitz, Wolter, Blaschke, Weber,
  Lichter, Ichimura, Collins, and Reifenberger]{cdkn2ameningiomas}
J~Bostrom, B~Meyer-Puttlitz, M~Wolter, B~Blaschke, RG~Weber, P~Lichter,
  K~Ichimura, VP~Collins, and G~Reifenberger.
\newblock Alterations of the tumor suppressor genes cdkn2a (p16(ink4a)),
  p14(arf), cdkn2b (p15(ink4b)), and cdkn2c (p18(ink4c)) in atypical and
  anaplastic meningiomas.
\newblock \emph{Am J Pathol}, 159\penalty0 (2):\penalty0 661–669, 2001.

\bibitem[Cabral et~al.(2018)Cabral, da~Gra{\c c}a Derengowski~Fonseca, and
  Mota]{Cabral2018-mz}
Bernardo~Pereira Cabral, Maria da~Gra{\c c}a Derengowski~Fonseca, and
  Fabio~Batista Mota.
\newblock The recent landscape of cancer research worldwide: a bibliometric and
  network analysis.
\newblock \emph{Oncotarget}, 9\penalty0 (55):\penalty0 30474--30484, July 2018.

\bibitem[Dud{\'a}s et~al.(2018)Dud{\'a}s, Dietl, Romani, Reinold, Glueckert,
  Schrott-Fischer, Dejaco, Johnson~Chacko, Tuertscher, Schartinger,
  et~al.]{dudas2018nerve}
J{\'o}zsef Dud{\'a}s, Wolfgang Dietl, Angela Romani, Susanne Reinold, Rudolf
  Glueckert, Anneliese Schrott-Fischer, Daniel Dejaco, Lejo Johnson~Chacko,
  Raphaela Tuertscher, Volker~Hans Schartinger, et~al.
\newblock Nerve growth factor (ngf)—receptor survival axis in head and neck
  squamous cell carcinoma.
\newblock \emph{International journal of molecular sciences}, 19\penalty0
  (6):\penalty0 1771, 2018.

\bibitem[Elmore et~al.(2021)Elmore, Greer, Daniels, Saxe, Melner, Krawiec,
  Cance, and Phelps]{Elmore2021-ft}
Lynne~W Elmore, Susanna~F Greer, Elvan~C Daniels, Charles~C Saxe, Michael~H
  Melner, Ginger~M Krawiec, William~G Cance, and William~C Phelps.
\newblock Blueprint for cancer research: Critical gaps and opportunities.
\newblock \emph{CA Cancer J. Clin.}, 71\penalty0 (2):\penalty0 107--139, March
  2021.

\bibitem[Franklin et~al.(2021)Franklin, Chari, Foreman, Seneviratne, Gruen,
  McCusker, Das, and Mcguinness]{knowledge-extraction-cohort-characteristics}
Jay Franklin, Shruthi Chari, Morgan Foreman, Oshani Seneviratne, Daniel Gruen,
  Jamie McCusker, Amar Das, and Deborah Mcguinness.
\newblock Knowledge extraction of cohort characteristics in research
  publications.
\newblock \emph{AMIA ... Annual Symposium proceedings. AMIA Symposium},
  2020:\penalty0 462--471, 01 2021.

\bibitem[Ganci et~al.(2020)Ganci, Pulito, Valsoni, Sacconi, Turco, Vahabi,
  Manciocco, Mazza, Meens, Karamboulas, et~al.]{ganci2020pi3k}
Federica Ganci, Claudio Pulito, Sara Valsoni, Andrea Sacconi, Chiara Turco,
  Mahrou Vahabi, Valentina Manciocco, Emilia Maria~Cristina Mazza, Jalna Meens,
  Christina Karamboulas, et~al.
\newblock Pi3k inhibitors curtail myc-dependent mutant p53 gain-of-function in
  head and neck squamous cell carcinomamyc mediates mutant p53 gof in hnscc.
\newblock \emph{Clinical Cancer Research}, 26\penalty0 (12):\penalty0
  2956--2971, 2020.

\bibitem[Hoischen et~al.(2008)Hoischen, Ehrler, Fassunke, Simon, Baudis,
  Landwehr, Radlwimmer, Lichter, Schramm, Becker, and Weber]{hoischen_cgh_2008}
A~Hoischen, M~Ehrler, J~Fassunke, M~Simon, M~Baudis, C~Landwehr, B~Radlwimmer,
  P~Lichter, J~Schramm, AJ~Becker, and RG~Weber.
\newblock Comprehensive characterization of genomic aberrations in
  gangliogliomas by cgh, array-based cgh and interphase fish.
\newblock \emph{Brain Pathol}, 18\penalty0 (3):\penalty0 326–337, 2008.

\bibitem[Huang et~al.(2021)Huang, Carrio-Cordo, Gao, Paloots, and
  Baudis]{progenetix2021}
Qingyao Huang, Paula Carrio-Cordo, Bo~Gao, Rahel Paloots, and Michael Baudis.
\newblock {The Progenetix oncogenomic resource in 2021}.
\newblock \emph{Database}, 2021, 07 2021.
\newblock ISSN 1758-0463.
\newblock \doi{10.1093/database/baab043}.
\newblock URL \url{https://doi.org/10.1093/database/baab043}.
\newblock baab043.

\bibitem[Kolluru et~al.(2020{\natexlab{a}})Kolluru, Adlakha, Aggarwal, Mausam,
  and Chakrabarti]{openie6}
Keshav Kolluru, Vaibhav Adlakha, Samarth Aggarwal, Mausam, and Soumen
  Chakrabarti.
\newblock Openie6: Iterative grid labeling and coordination analysis for open
  information extraction, 2020{\natexlab{a}}.

\bibitem[Kolluru et~al.(2020{\natexlab{b}})Kolluru, Aggarwal, Rathore,
  Chakrabarti, et~al.]{imojie}
Keshav Kolluru, Samarth Aggarwal, Vipul Rathore, Soumen Chakrabarti, et~al.
\newblock Imojie: Iterative memory-based joint open information extraction.
\newblock \emph{arXiv preprint arXiv:2005.08178}, 2020{\natexlab{b}}.

\bibitem[Landolsi et~al.(2022)Landolsi, Hlaoua, Ben, and
  Romdhane]{10.1007/s10115-022-01779-1}
Mohamed~Yassine Landolsi, Lobna Hlaoua, Ben, and Lotfi Romdhane.
\newblock Information extraction from electronic medical documents: State of
  the art and future research directions.
\newblock \emph{Knowl. Inf. Syst.}, 65\penalty0 (2):\penalty0 463–516, nov
  2022.
\newblock ISSN 0219-1377.
\newblock \doi{10.1007/s10115-022-01779-1}.
\newblock URL \url{https://doi.org/10.1007/s10115-022-01779-1}.

\bibitem[Lassmann et~al.(2007)Lassmann, Weis, Makowiec, Roth, Danciu, Hopt, and
  Werner]{lassmanCNVinCRC2007}
S~Lassmann, R~Weis, F~Makowiec, J~Roth, M~Danciu, U~Hopt, and M~Werner.
\newblock Array cgh identifies distinct dna copy number profiles of oncogenes
  and tumor suppressor genes in chromosomal- and microsatellite-unstable
  sporadic colorectal carcinomas.
\newblock \emph{J Mol Med (Berl)}, 85\penalty0 (3):\penalty0 293–304, 2007.

\bibitem[Lee et~al.(2019)Lee, Parameswaran, Sandoval-Schaefer, Eoh, Yang, Zhu,
  Mehra, Sharma, Gaffney, Perry, et~al.]{lee2019combined}
Jong~Woo Lee, Janaki Parameswaran, Teresa Sandoval-Schaefer, Kyung~Jin Eoh,
  Dong-hua Yang, Fang Zhu, Ranee Mehra, Roshan Sharma, Stephen~G Gaffney,
  Elizabeth~B Perry, et~al.
\newblock Combined aurora kinase a (aurka) and wee1 inhibition demonstrates
  synergistic antitumor effect in squamous cell carcinoma of the head and
  neckcombined aurka and wee1 inhibition in hnscc.
\newblock \emph{Clinical Cancer Research}, 25\penalty0 (11):\penalty0
  3430--3442, 2019.

\bibitem[Liu et~al.(2016)Liu, Chen, Jagannatha, and Yu]{liu2016learning}
Feifan Liu, Jinying Chen, Abhyuday Jagannatha, and Hong Yu.
\newblock Learning for biomedical information extraction: Methodological review
  of recent advances, 2016.

\bibitem[Lubowitz et~al.(2021)Lubowitz, Brand, and Rossi]{LUBOWITZ20213221}
James~H. Lubowitz, Jefferson~C. Brand, and Michael~J. Rossi.
\newblock Medical journal content continues rapid growth.
\newblock \emph{Arthroscopy: The Journal of Arthroscopic and Related Surgery},
  37\penalty0 (11):\penalty0 3221--3222, 2021.
\newblock ISSN 0749-8063.
\newblock \doi{https://doi.org/10.1016/j.arthro.2021.09.006}.
\newblock URL
  \url{https://www.sciencedirect.com/science/article/pii/S0749806321008227}.

\bibitem[Luo et~al.(2022)Luo, Lai, Wei, Arighi, and Lu]{biored}
Ling Luo, Po-Ting Lai, Chih-Hsuan Wei, Cecilia~N Arighi, and Zhiyong Lu.
\newblock Biored: a rich biomedical relation extraction dataset.
\newblock \emph{Briefings in Bioinformatics}, 23\penalty0 (5):\penalty0
  bbac282, 2022.

\bibitem[Macnee et~al.(2021)Macnee, Pérez-Palma, Schumacher-Bass, Dalton, Leu,
  Blankenberg, and Lal]{10.1093/bioinformatics/btab365}
Marie Macnee, Eduardo Pérez-Palma, Sarah Schumacher-Bass, Jarrod Dalton,
  Costin Leu, Daniel Blankenberg, and Dennis Lal.
\newblock {SimText: a text mining framework for interactive analysis and
  visualization of similarities among biomedical entities}.
\newblock \emph{Bioinformatics}, 37\penalty0 (22):\penalty0 4285--4287, 05
  2021.
\newblock ISSN 1367-4803.
\newblock \doi{10.1093/bioinformatics/btab365}.
\newblock URL \url{https://doi.org/10.1093/bioinformatics/btab365}.

\bibitem[Mirabelli et~al.(2019)Mirabelli, Coppola, and
  Salvatore]{Mirabelli2019-wv}
Peppino Mirabelli, Luigi Coppola, and Marco Salvatore.
\newblock Cancer cell lines are useful model systems for medical research.
\newblock \emph{Cancers (Basel)}, 11\penalty0 (8):\penalty0 1098, August 2019.

\bibitem[Mohamed et~al.(2019)Mohamed, Nováček, and
  Nounu]{10.1093/bioinformatics/btz600}
Sameh~K Mohamed, Vít Nováček, and Aayah Nounu.
\newblock {Discovering protein drug targets using knowledge graph embeddings}.
\newblock \emph{Bioinformatics}, 36\penalty0 (2):\penalty0 603--610, 08 2019.
\newblock ISSN 1367-4803.
\newblock \doi{10.1093/bioinformatics/btz600}.
\newblock URL \url{https://doi.org/10.1093/bioinformatics/btz600}.

\bibitem[Mungall et~al.(2012)Mungall, Torniai, Gkoutos, Lewis, and
  Haendel]{uberon}
Christopher~J. Mungall, Carlo Torniai, Georgios~V. Gkoutos, Suzanna~E. Lewis,
  and Melissa~A. Haendel.
\newblock Uberon, an integrative multi-species anatomy ontology.
\newblock \emph{Genome Biology}, 13\penalty0 (1):\penalty0 R5, Jan 2012.
\newblock ISSN 1474-760X.
\newblock \doi{10.1186/gb-2012-13-1-r5}.
\newblock URL \url{https://doi.org/10.1186/gb-2012-13-1-r5}.

\bibitem[Patra et~al.(2017)Patra, Young, Llewellyn, Senapati, and
  Mathew]{patra2017braf}
Satyajit Patra, Vanesa Young, Leslie Llewellyn, Jitendra~N Senapati, and Jesil
  Mathew.
\newblock Braf, kras and pik3ca mutation and sensitivity to trastuzumab in
  breast cancer cell line model.
\newblock \emph{Asian Pacific journal of cancer prevention: APJCP}, 18\penalty0
  (8):\penalty0 2209, 2017.

\bibitem[Patrick et~al.(2011)Patrick, Nguyen, Wang, and
  Li]{knowledge-discovery-reuse-pipeline}
Jon Patrick, Hoang Nguyen, Yefeng Wang, and Min Li.
\newblock A knowledge discovery and reuse pipeline for information extraction
  in clinical notes.
\newblock \emph{Journal of the American Medical Informatics Association :
  JAMIA}, 18:\penalty0 574--9, 09 2011.
\newblock \doi{10.1136/amiajnl-2011-000302}.

\bibitem[Qu and Cui(2021)]{10.1093/bioinformatics/btab851}
Jialin Qu and Yuehua Cui.
\newblock {Gene set analysis with graph-embedded kernel association test}.
\newblock \emph{Bioinformatics}, 38\penalty0 (6):\penalty0 1560--1567, 12 2021.
\newblock ISSN 1367-4803.
\newblock \doi{10.1093/bioinformatics/btab851}.
\newblock URL \url{https://doi.org/10.1093/bioinformatics/btab851}.

\bibitem[Quimbaya et~al.(2016)Quimbaya, M{\'u}nera, Rivera, Rodr{\'\i}guez,
  Velandia, Pe{\~n}a, and Labb{\'e}]{quimbaya2016named}
Alexandra~Pomares Quimbaya, Alejandro~Sierra M{\'u}nera, Rafael
  Andr{\'e}s~Gonz{\'a}lez Rivera, Juli{\'a}n Camilo~Daza Rodr{\'\i}guez, Oscar
  Mauricio~Mu{\~n}oz Velandia, Angel Alberto~Garcia Pe{\~n}a, and Cyril
  Labb{\'e}.
\newblock Named entity recognition over electronic health records through a
  combined dictionary-based approach.
\newblock \emph{Procedia Computer Science}, 100:\penalty0 55--61, 2016.

\bibitem[Rao et~al.(2010)Rao, Edwards, Joshi, Siu, and
  Riggins]{raoGlioblastomas2008}
SK~Rao, J~Edwards, AD~Joshi, IM~Siu, and GJ~Riggins.
\newblock A survey of glioblastoma genomic amplifications and deletions.
\newblock \emph{J Neurooncol}, 96\penalty0 (2):\penalty0 169–179, 2010.

\bibitem[Santra et~al.(2017)Santra, Roche, Conlon, O’Donovan, Crown,
  O’Connor, and Kolch]{santra2017identification}
Tapesh Santra, Sandra Roche, Neil Conlon, Norma O’Donovan, John Crown, Robert
  O’Connor, and Walter Kolch.
\newblock Identification of potential new treatment response markers and
  therapeutic targets using a gaussian process-based method in lapatinib
  insensitive breast cancer models.
\newblock \emph{Plos one}, 12\penalty0 (5):\penalty0 e0177058, 2017.

\bibitem[Siegel et~al.(2017)Siegel, Miller, and Jemal]{seer2017}
Rebecca~L. Siegel, Kimberly~D. Miller, and Ahmedin Jemal.
\newblock Cancer statistics, 2017.
\newblock \emph{CA: A Cancer Journal for Clinicians}, 67\penalty0 (1):\penalty0
  7--30, 2017.
\newblock \doi{https://doi.org/10.3322/caac.21387}.
\newblock URL
  \url{https://acsjournals.onlinelibrary.wiley.com/doi/abs/10.3322/caac.21387}.

\bibitem[Siegel et~al.(2022)Siegel, Miller, Fuchs, and Jemal]{Siegel2022-rb}
Rebecca~L Siegel, Kimberly~D Miller, Hannah~E Fuchs, and Ahmedin Jemal.
\newblock Cancer statistics, 2022.
\newblock \emph{CA Cancer J. Clin.}, 72\penalty0 (1):\penalty0 7--33, January
  2022.

\bibitem[Singh et~al.(2011)Singh, Odriozola, Guan, Kennedy, and
  Chan]{singh2011characterization}
Gobind Singh, Leticia Odriozola, Hong Guan, Colin~R Kennedy, and Andrew~M Chan.
\newblock Characterization of a novel pten mutation in mda-mb-453 breast
  carcinoma cell line.
\newblock \emph{BMC cancer}, 11:\penalty0 1--11, 2011.

\bibitem[Sioutos et~al.(2007)Sioutos, Coronado, Haber, Hartel, Shaiu, and
  Wright]{ncithesaurus}
Nicholas Sioutos, Sherri~de Coronado, Margaret~W. Haber, Frank~W. Hartel,
  Wen-Ling Shaiu, and Lawrence~W. Wright.
\newblock Nci thesaurus: A semantic model integrating cancer-related clinical
  and molecular information.
\newblock \emph{J. of Biomedical Informatics}, 40\penalty0 (1):\penalty0
  30–43, feb 2007.
\newblock ISSN 1532-0464.
\newblock \doi{10.1016/j.jbi.2006.02.013}.
\newblock URL \url{https://doi.org/10.1016/j.jbi.2006.02.013}.

\bibitem[Smith et~al.(2022)Smith, Papadopoulos, Braschler, and
  Stockinger]{SMITH2022101938}
Ellery Smith, Dimitris Papadopoulos, Martin Braschler, and Kurt Stockinger.
\newblock Lillie: Information extraction and database integration using
  linguistics and learning-based algorithms.
\newblock \emph{Information Systems}, 105:\penalty0 101938, 2022.
\newblock ISSN 0306-4379.
\newblock \doi{https://doi.org/10.1016/j.is.2021.101938}.
\newblock URL
  \url{https://www.sciencedirect.com/science/article/pii/S030643792100137X}.

\bibitem[Subramanian et~al.(2020)Subramanian, Baldini, Ravichandran,
  Katz-Rogozhnikov, Natesan~Ramamurthy, Sattigeri, Varshney, Wang, Mangalath,
  and Kleiman]{Subramanian}
Shivashankar Subramanian, Ioana Baldini, Sushma Ravichandran, Dmitriy~A.
  Katz-Rogozhnikov, Karthikeyan Natesan~Ramamurthy, Prasanna Sattigeri, Kush~R.
  Varshney, Annmarie Wang, Pradeep Mangalath, and Laura~B. Kleiman.
\newblock A natural language processing system for extracting evidence of drug
  repurposing from scientific publications.
\newblock \emph{Proceedings of the AAAI Conference on Artificial Intelligence},
  34\penalty0 (08):\penalty0 13369--13381, Apr. 2020.
\newblock \doi{10.1609/aaai.v34i08.7052}.
\newblock URL \url{https://ojs.aaai.org/index.php/AAAI/article/view/7052}.

\bibitem[Tweedie et~al.(2021)Tweedie, Braschi, Gray, Jones, Seal, Yates, and
  Bruford]{hugogene}
Susan Tweedie, Bryony Braschi, Kristian Gray, Tamsin E~M Jones, Ruth~L Seal,
  Bethan Yates, and Elspeth~A Bruford.
\newblock {Genenames.Org}: The {HGNC} and {VGNC} resources in 2021.
\newblock \emph{Nucleic Acids Res.}, 49\penalty0 (D1):\penalty0 D939--D946,
  January 2021.

\bibitem[Vogelstein et~al.(2013)Vogelstein, Papadopoulos, Velculescu, Zhou,
  Diaz, and Kinzler]{vogelstein2013cance}
B~Vogelstein, N~Papadopoulos, VE~Velculescu, S~Zhou, LA~Diaz, and KW~Kinzler.
\newblock Cancer genome landscapes.
\newblock \emph{Science}, 339\penalty0 (6127):\penalty0 1546–1558, 2013.

\bibitem[Xu et~al.(2020)Xu, Kim, Song, Jeong, Kim, Kang, Rousseau, Li, Xu,
  Torvik, Bu, Chen, Ebeid, Li, and Ding]{71be196e376b4fa1a71a1bbda69a8d4e}
Jian Xu, Sunkyu Kim, Min Song, Minbyul Jeong, Donghyeon Kim, Jaewoo Kang,
  Justin~F. Rousseau, Xin Li, Weijia Xu, Vetle~I. Torvik, Yi~Bu, Chongyan Chen,
  Islam~Akef Ebeid, Daifeng Li, and Ying Ding.
\newblock Building a pubmed knowledge graph.
\newblock \emph{Scientific data}, 7\penalty0 (1), December 2020.
\newblock ISSN 2052-4463.
\newblock \doi{10.1038/s41597-020-0543-2}.

\end{thebibliography}
